\documentclass[10pt, a4paper]{article}
\usepackage{lrec2022} % this is the new LREC2022 Style
\usepackage{multibib}
\newcites{languageresource}{Language Resources}
\usepackage{graphicx}
\usepackage{tabularx}
\usepackage{soul}
\usepackage{titlesec}
\titleformat{\section}{\normalfont\large\bfseries\center}{\thesection.}{1em}{}
\titleformat{\subsection}{\normalfont\SmallTitleFont\bfseries\raggedright}{\thesubsection.}{1em}{}
\titleformat{\subsubsection}{\normalfont\normalsize\bfseries\raggedright}{\thesubsubsection.}{1em}{}
\renewcommand\thesection{\arabic{section}}
\renewcommand\thesubsection{\thesection.\arabic{subsection}}
\renewcommand\thesubsubsection{\thesubsection.\arabic{subsubsection}}
\usepackage{epstopdf}
\usepackage[utf8]{inputenc}

\usepackage{hyperref}
\usepackage{xstring}

\usepackage{booktabs}
\usepackage{multirow}
\usepackage{graphicx}

\usepackage{color}

\usepackage{pgfplots}

\usepackage{microtype}
\usepackage{xcolor}
\usepackage{amsmath,amssymb}

\title{XLM-T: Multilingual Language Models in Twitter \\ for Sentiment Analysis and Beyond}%%\\ \vspace*{.5\baselineskip} \normalfont{ The Title \ul{Must Be} Capitalised as in:\\ \vspace*{.5\baselineskip} \textbf{The Rise and Fall of Ziggy Stardust and the Spiders from Mars}}}

\name{Francesco Barbieri$^{\clubsuit}$, Luis Espinosa Anke$^{\diamondsuit}$, Jose Camacho-Collados$^{\diamondsuit}$} 

\address{$^{\clubsuit}$ Snap Inc., $^{\diamondsuit}$ Cardiff NLP, School of Computer Science and Informatics, Cardiff University \\
         $^{\clubsuit}$ Santa Monica, California, USA  $^{\diamondsuit}$ Cardiff, Wales, United Kingdom \\
         fbarbieri@snap.com,
         \{espinosa-ankel,camachocolladosj\}@cardiff.ac.uk\\}

\abstract{
Language models are ubiquitous in current NLP, and their multilingual capacity has recently attracted considerable attention. However, current analyses have almost exclusively focused on (multilingual variants of) standard benchmarks, and have relied on clean pre-training and task-specific corpora as multilingual signals. In this paper, we introduce XLM-T, a model to train and evaluate multilingual language models in Twitter. In this paper we provide: (1) a new strong multilingual baseline consisting of an XLM-R \cite{conneau-etal-2020-unsupervised} model pre-trained on millions of tweets in over thirty languages, alongside starter code to subsequently fine-tune on a target task; and (2) a set of unified sentiment analysis Twitter datasets in eight different languages and a XLM-T model fine-tuned on them. %\cite{barbieri-etal-2020-tweeteval}. 
 \\ \newline  }

\begin{document}

\maketitleabstract

\section{Introduction}

Multilingual NLP is increasingly becoming popular. Despite the concerning disparity in terms of language resource availability \cite{joshi-etal-2020-state}, the advent of Language Models (LMs) has indisputably enabled a myriad of multilingual architectures to fluorish, ranging from LSTMs to the arguably more popular transformer-based models \cite{chronopoulou2019embarrassingly,pires2019multilingual}. Multilingual LMs integrate streams of multilingual textual data without being tied to one single task, learning \textit{general-purpose multilingual representations} \cite{hu2020xtreme}. As testimony of this landscape, we find multilingual variants stemming from well-known monolingual LMs, which have now become a standard among the NLP community. For instance, mBERT from BERT \cite{devlin-etal-2019-bert}, mT5 \cite{xue2020mt5} from T5 \cite{raffel2020exploring} or XLM-R \cite{conneau-etal-2020-unsupervised} from RoBERTa \cite{liu2019roberta}. Social media data, however, and specifically Twitter (the platform we focus on in this paper), seem to be so far surprisingly neglected from this trend of massive multilingual pretraining. This may be due to, in addition to its well-known uncurated nature \cite{derczynski2013twitter}, because of discoursive and platform-specific factors such as out-of-distribution samples, misspellings, slang, vulgarisms, emoji and multimodality, among others \cite{barbieri2018semeval,camacho2020learning}. This is an important consideration, as there is ample agreement that the quality of LM-based multilingual representations is strongly correlated with typological similarity \cite{hu2020xtreme}, which is somewhat blurred out in the context of Twitter.

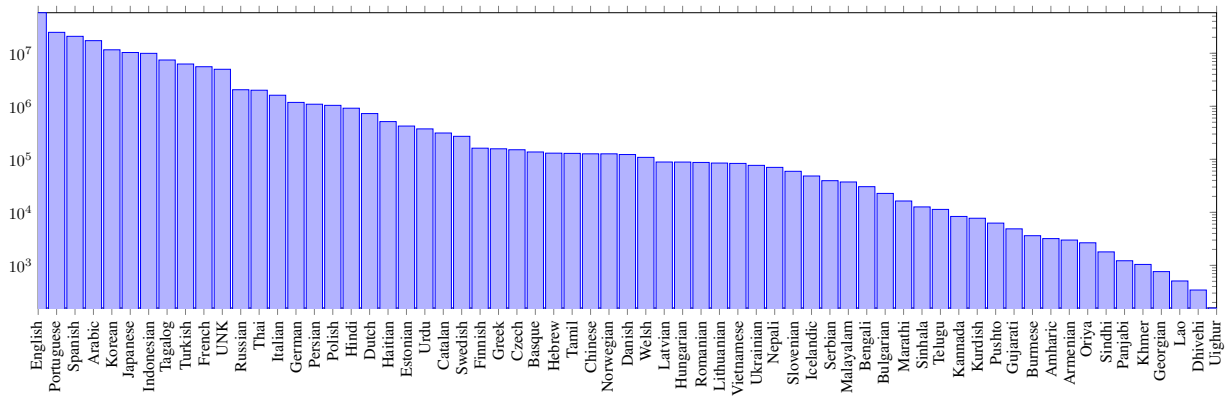
\begin{figure*}[t] 
\vspace{-1.8cm}
\begin{tikzpicture}
\centering
%\small
\hspace{-0.38cm}
%\resizebox{\textwidth}{!}{
\scalebox{0.61}{ 
\begin{axis}[
    width=27cm,
    height=8cm,
    enlargelimits=false,
    %width=0.8\textwidth,
    ybar,
        enlarge x limits=false,
        %ybar interval,
        %xbar interval,
        x tick label as interval=false,
        x filter/.code=\pgfmathparse{#1-0.5},
        xtick={},
        xmajorgrids=false,
    ymode=log,
    %log ticks with fixed point,
    %xlabel={Language},
    symbolic x coords={English, Portuguese, Spanish, Arabic, Korean, Japanese, Indonesian, Tagalog, Turkish, French, UNK, Russian, Thai, Italian, German, Persian, Polish, Hindi, Dutch, Haitian, Estonian, Urdu, Catalan, Swedish, Finnish, Greek, Czech, Basque, Hebrew, Tamil, Chinese, Norwegian, Danish, Welsh, Latvian, Hungarian, Romanian, Lithuanian, Vietnamese, Ukrainian, Nepali, Slovenian, Icelandic, Serbian, Malayalam, Bengali, Bulgarian, Marathi, Sinhala, Telugu, Kannada, Kurdish, Pushto, Gujarati, Burmese, Amharic, Armenian, Oriya, Sindhi, Panjabi, Khmer, Georgian, Lao, Dhivehi, Uighur},
    xtick=data,
    xticklabel style={rotate=90},
    nodes near coords align={vertical},
]
    \addplot coordinates {(English,58213411) (Portuguese,24820942) (Spanish,20881837) (Arabic,17304714) (Korean,11632987) (Japanese,10347324) (Indonesian,9972362) (Tagalog,7470110) (Turkish,6275754) (French,5578560) (UNK,4991864) (Russian,2053497) (Thai,2010622) (Italian,1617315) (German,1184639) (Persian,1095121) (Polish,1042317) (Hindi,920833) (Dutch,729846) (Haitian,514719) (Estonian,424054) (Urdu,374955) (Catalan,313657) (Swedish,271589) (Finnish,162245) (Greek,157963) (Czech,151565) (Basque,137586) (Hebrew,130727) (Tamil,129511) (Chinese,127003) (Norwegian,126981) (Danish,123160) (Welsh,108898) (Latvian,89005) (Hungarian,88791) (Romanian,87369) (Lithuanian,85034) (Vietnamese,83744) (Ukrainian,76960) (Nepali,70510) (Slovenian,59465) (Icelandic,48481) (Serbian,39561) (Malayalam,37402) (Bengali,30503) (Bulgarian,22765) (Marathi,16375) (Sinhala,12664) (Telugu,11365) (Kannada,8371) (Kurdish,7745) (Pushto,6268) (Gujarati,4885) (Burmese,3629) (Amharic,3200) (Armenian,3003) (Oriya,2666) (Sindhi,1796) (Panjabi,1222) (Khmer,1042) (Georgian,763) (Lao,508) (Dhivehi,343) (Uighur,157)};
\end{axis}
}
\end{tikzpicture}
\vspace{-1.5cm}
\caption{\label{fig:unlabelled} Distribution of languages of the 198M tweets used to finetune the Twitter-based language model (log scale). UNK corresponds to unidentified tweets according to the Twitter API.}
\end{figure*}

In this paper, we bridge this gap by introducing a toolkit for evaluating multilingual Twitter-specific language models. This framework, which we make available to the NLP community, is initially comprised of a large multilingual Twitter-specific LM based on XLM-R checkpoints (Section \ref{sec:language_model_socialmedia}), from which we report an initial set of baseline results in different settings (including zero-shot). Moreover, we provide starter code for analyzing, fine-tuning and evaluating existing language models. To carry out a comprehensive multilingual evaluation, while also laying the foundations for future extensions, we devise a unified dataset in 8 languages for sentiment analysis (which we call \textit{Unified Multilingual Sentiment Analysis Benchmark}, UMSAB henceforth), as this task is by far the most studied problem in NLP in Twitter (cf., e.g., \cite{salameh2015,zhou2016,meng2012,chen2018adversarial,Rasooli2018,vilares2017,barnes-etal-2019-sentiment,patwa-etal-2020-semeval,barriere-balahur-2020-improving}). XLM-T and associated data is released at \url{https://github.com/cardiffnlp/xlm-t}.

Finally, in order to have a solid point of comparison with respect to standard English Twitter tasks, we also report results on the TweetEval framework \cite{barbieri-etal-2020-tweeteval}. Our results suggest that when fine-tuning task-specific Twitter-based multilingual LMs, a domain-specific model proves more consistent than its general-domain counterpart, and that in some cases a smart selection of training data may be preferred than large-scale fine-tuning on many languages.

\section{XLM-T: Language Models in Twitter}
\label{sec:language_model_socialmedia}

Our framework revolves around Twitter-specific language models. In particular, we train our own multilingual language-specific language model (Section \ref{sec:language_model_pret}), which we then fine-tune for various monolingual and multilingual applications, and for which we provide a suitable interface (Section \ref{sec:language_model_finetune}). Additionally, we complement these functionalities with starter code for these and other typical Twitter-related NLP tasks (Section \ref{sec:startercode}), e.g., computing tweet embeddings and multilingual sentiment analysis evaluation. % based on the released language models. %(Section \ref{sec:addcode}).

\subsection{Released Language Models}
\label{sec:language_model_pret}

%\subsection{Multilingual Twitter-based Pretraining}
%\label{sec:multlanguage_model}
%\section{Developing Multilingual Resources}

%In this section we briefly cover the process of data gathering for pretraining as well as the design choices behind the compilation and curation of our newly proposed dataset for multilingual SA.

%\paragraph{Multilingual Twitter-based XLM-R}
%\label{sec:language_model}

We used the Twitter API to retrieve 198M tweets\footnote{1,724 million tokens (12G of uncompressed text).} posted between May'18 and March'20, which are our source data for LM pretraining.
%collected a large corpus of tweets to be fed for training a LM
%In order to train train a large language model on Twitter we use an unlabelled corpus, that includes 193M tweets\footnote{1,724 million tokens (12G of uncompressed text).} that was crawled with the stream API from May'18 to March'20. 
We only considered tweets with at least three tokens and with no URLs to avoid bot tweets and spam advertising. Additionally, we did not perform language filtering, aiming at capturing a general distribution. Figure~\ref{fig:unlabelled} lists the 30 most represented languages by frequency, showing a prevalence of widely spoken languages such as English, Portuguese and Spanish, with the first significant drop in frequency affecting Russian at the 11th position.

In terms of opting for pretraining a LM from scratch or building upon an existing one, we follow \cite{gururangan-etal-2020-dont} and \cite{barbieri-etal-2020-tweeteval} and \textit{continue training} an XLM-R language model from publicly available checkpoints\footnote{\url{https://huggingface.co/xlm-roberta-base}.}, which we selected due to the high results it has achieved in several multilingual NLP tasks \cite{hu2020xtreme}. We use the same masked LM objective, and train until convergence in a validation set. The model converged after about 14 days on 8 NVIDIA V100 GPUs.\footnote{The estimated cost for the language model pre-training is USD 5,000 on Google Cloud.}

While this multilingual language model (referred to as \textit{XLM-Twitter} henceforth) is the main focus on this paper, our toolkit also integrates monolingual language models of any nature, including the English monolingual Twitter models released in \newcite{barbieri-etal-2020-tweeteval} and \newcite{nguyen-etal-2020-bertweet}.

\subsection{Language Model Fine-tuning}
\label{sec:language_model_finetune}

In this section we explain the fine-tuning implementation of our framework. The main task evaluated in this paper is tweet classification, for which we provide unified datasets. One of the main differences with respect to standard fine-tuning is that we integrate the adapter technique \cite{houlsby2019parameter}, by means of which we freeze the LM and only fine-tune one additional classification layer. We follow the same adapter configuration proposed in \newcite{pfeiffer2020adapterfusion}. This technique provides benefits in terms of memory and speed, which in practice facilitates the usage of multilingual language models for a wider set of NLP practitioners and researchers.%\footnote{In this paper we do not focus on evaluating the reliability of the adapter technique as this was tested in previous papers. Instead, we add this functionality as default for the code and all our experiments.} \luis{Yo quitaría esta footnote.}

%\red{Show snapshots of code on how to use the language models for fine-tuning.}

%\subsubsection{Model Fine-tuning}

%\subsubsection{Multilingual Fine-tuning}

%\subsection{Software Interface}
%\label{sec:addcode}

\subsection{Starter code} 
\label{sec:startercode}
In order to enable fast prototyping on our framework, in addition to datasets and pretrained models we also provide Python code for feature extraction from Tweets (i.e., obtaining tweet embeddings), tweet classification, model fine-tuning,  %\luis{I'd remove ref to adapter here} which again, is based on the adapter technique (cf. Section \ref{cross:eval}),
and evaluation.%\footnote{A short video showcasing our repository is available at \url{www.tinyurl.com/demo-xlmt}.}

\begin{figure}[t]
\includegraphics[width=\columnwidth]{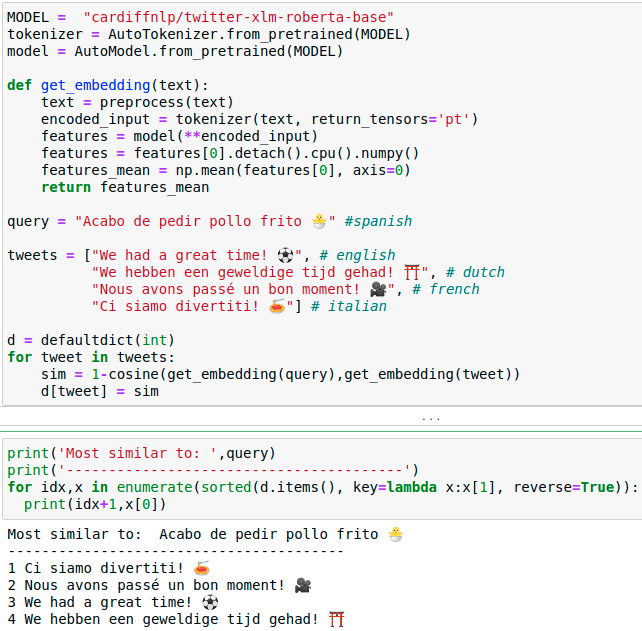}
\caption{\label{fig:code} Code snippet showcasing the feature extraction and tweet similarity interface. Note that using our Twitter-specific XLM-R model leads to emoji playing a crucial role in the semantics of the tweet.}
\end{figure}

%\begin{itemize}
\paragraph{Feature extraction.} Figure \ref{fig:code} shows sample code on how to extract tweet embeddings using our XLM-T language model, including its applicability for tweet similarity.
    
\begin{figure}[t]
\includegraphics[width=\columnwidth]{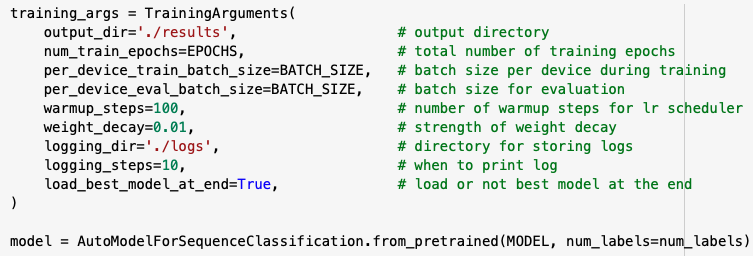}
\includegraphics[width=\columnwidth]{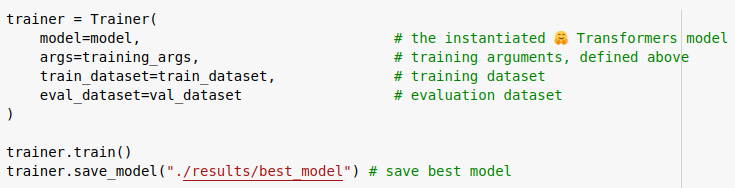}
\caption{\label{fig:finetuning} Fine-tuning procedure including the declaration of dataset and parameters, training procedure and saving of the model.}
\end{figure}

\paragraph{Fine-tuning.} Figure \ref{fig:finetuning} shows the fine-tuning procedure using a custom language model. This process can be performed with either adapters (used in our evaluation for efficiency) or the more standard language model fine-tuning. In practice, note that both options would be implemented in a very similar way, as both sit on top of the Huggingface \texttt{transformers} library.
    
\begin{figure}[t]
\includegraphics[width=\columnwidth]{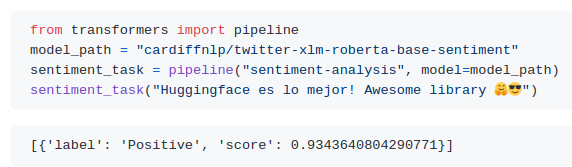}
\caption{\label{fig:codeinference} Sentiment analysis inference using XLM-T.}
\end{figure}

\paragraph{Inference (tweet classification).} We provide an easy interface to perform inference with our fine-tuned models. To this end, we rely on Hugging Face's \textit{pipelines}. Figure \ref{fig:codeinference} shows an example for a sentiment prediction using our XLM-T model fine-tuned on UMSAB. Note that, while the examples provided are for sentiment analysis, any tweet classification task such as those included in TweetEval are compatible.
    
\paragraph{Evaluation.} Finally, XLM-T includes evaluation code to seamlessly evaluate any language model on sentiment analysis, either focusing on a subset or all of the languages included in UMSAB (cf. Section \ref{cross:eval}). Specifically, we provide bash scripts which handle input arguments such as gold test data, prediction files and target language(s). % bash code for evaluation in any of the languages included in our framework, with flexible arguments involving the gold test data, predictions and language.
    
%\end{itemize}

\section{Evaluation}

We assess the reliability of our released multilingual Twitter-specific language model in three different ways: (1) we perform an evaluation on a wide range of English-specific datasets (Section \ref{mono:eval}); (2) we compose a large multilingual benchmark for sentiment analysis where we assess the multilingual capabilities of the language model (Section~\ref{cross:eval}); (3) we perform a qualitative analysis based on cross-lingual tweet similarity (Section \ref{qualitative}).

\paragraph{Experimental Setting. }
\label{sec:exp_setting}

\begin{table*} %[bp]
\renewcommand{\arraystretch}{1.25}
\setlength{\tabcolsep}{2.0pt}
\resizebox{\textwidth}{!}{
\begin{tabular}{c|c|c|c|c|c|c|c||c}
\toprule
   &
  \textbf{Emoji} &
  \textbf{Emotion} &
  \textbf{Hate} &
  \textbf{Irony} &
  \textbf{Offensive} &
  \textbf{Sentiment} &
  \textbf{Stance} &
  \textbf{ALL} \\ \hline \hline

%\multicolumn{1}{c|}{\multirow{9}{*}{Test}} &
  SVM &
  29.3 &
  64.7 &
  36.7 &
  61.7 &
  52.3 &
  62.9 &
  67.3 &
  53.5 \\
  FastText &
  25.8 &
  65.2 &
  50.6 &
  63.1 &
  73.4 &
  62.9 &
  65.4 &
 58.1 \\

  BLSTM &
  24.7 &
  66.0 &
  52.6 &
  62.8 &
  71.7 &
  58.3 &
  59.4 &
  56.5  \\
  RoB-Bs &
  30.9{\small $\pm$0.2} (30.8) &
  76.1{\small $\pm$0.5} (76.6) &
  46.6{\small $\pm$2.5} (44.9) &
  59.7{\small $\pm$5.0} (55.2) &
  79.5{\small $\pm$0.7} (78.7) &
  71.3{\small $\pm$1.1} (72.0) &
  68{\small $\pm$0.8} (70.9) &
  61.3 \\
  RoB-RT &
  31.4{\small $\pm$0.4} (\textbf{31.6}) &
  78.5{\small $\pm$1.2} (\textbf{79.8}) &
  52.3{\small $\pm$0.2} (\textbf{55.5}) &
  61.7{\small $\pm$0.6} (62.5) &
  80.5{\small $\pm$1.4} (\textbf{81.6}) &
  72.6{\small $\pm$0.4} (\textbf{72.9}) &
  69.3{\small $\pm$1.1} (\textbf{72.6}) &
  \textbf{65.2} \\
  RoB-Tw &
  29.3{\small $\pm$0.4} (29.5) &
  72.0{\small $\pm$0.9} (71.7) &
  46.9{\small $\pm$2.9} (45.1) &
  65.4{\small $\pm$3.1} (65.1) &
  77.1{\small $\pm$1.3} (78.6) &
  69.1{\small $\pm$1.2} (69.3) &
  66.7{\small $\pm$1.0} (67.9) &
  61.0 \\ \cline{1-9}

  XLM-R &
  28.6{\small $\pm$0.7} (27.7) &
  72.3{\small $\pm$3.6} (68.5) &
  44.4{\small $\pm$0.7} (43.9) &
  57.4{\small $\pm$4.7} (54.2) &
  75.7{\small $\pm$1.9} (73.6) &
  68.6{\small $\pm$1.2} (69.6) &
  65.4{\small $\pm$0.8} (66.0) &
  57.6 \\ 
  
  XLM-Tw &
  30.9{\small $\pm$0.5} (30.8) &
  77.0{\small $\pm$1.5} (78.3) &
  50.8{\small $\pm$0.6} (51.5) &
  69.9{\small $\pm$1.0} (\textbf{70.0}) &
  79.9{\small $\pm$0.8} (79.3) &
  72.3{\small $\pm$0.2} (72.3) &
  67.1{\small $\pm$1.4} (68.7) &
  64.4 \\ \cline{1-9}
  
%BERT-Tw 
SotA &
  \textit{33.4} &
  \textit{79.3} &
  \textit{56.4} &
  \textit{82.1} &
  \textit{79.5} &
  \textit{73.4} &
  \textit{71.2} &
 % 32.4{\small $\pm$0.9} (\textbf{33.4}) &
 % 79.6{\small $\pm$0.6} (79.3) &
 % 53.9{\small $\pm$2.6} (\textbf{56.4}) &
 % 80.3{\small $\pm$2.0} (\textbf{82.1}) &
 % 80.5{\small $\pm$0.9} (79.5) &
%  72.6{\small $\pm$1.0} (\textbf{73.4}) &
%  70.0{\small $\pm$1.1} (71.2) &
  %\textbf{67.9} \\ \cline{1-9} 
  %\textit{SotA} &
 % 36.0* &
 % - &
 % 65.1 &
 % 70.5 &
 % 82.9 &
 % 68.5 &
 % 71.0 &

  \textit{67.9} \\ \hline \hline
\multicolumn{1}{c|}{\textbf{Metric}} &
  M-F1 &
  M-F1 &
  M-F1 &
  F$^{(i)}$ &
  M-F1 &
  M-Rec &
  AVG (F$^{(a)}$,$F^{(f)}$) & TE
   \\ \bottomrule
\end{tabular}
}
\caption{\label{table-results} TweetEval test results. For neural models we report both the average result from three runs and its standard deviation, and the best result according to the validation set (parentheses). \textit{SotA} results correspond to the best TweetEval reported system, i.e., BERTweet.}
\end{table*}

%systems in the original shared tasks - they are included for completeness as they not directly comparable.  %as were reported under different settings and not in this unified benchmark, . 
%Splits might differ, and * indicates that a larger training set is used.
%Instead of fine-tuning whole LMs, we apply the adapter technique \cite{houlsby2019parameter}, by means of which we freeze the LM and only fine-tune one additional classification layer. We follow the same adapter configuration proposed in \newcite{pfeiffer2020adapterfusion}.
In each experiment we perform three runs with different seeds, and use early stopping on the validation loss. We only tune the learning rate (0.001 and 0.0001) and, unless noted otherwise, all results we report are the average of three runs of macro-average F1 scores.%\footnote{Standard deviation and best run results are provided, for completeness, in the supplementary material.}
In terms of models, we evaluate a standard pre-trained \textbf{XLM-R} and \textbf{XLM-Twitter}, our XLM-R model pretrained on a multilingual Twitter dataset starting from XLM-R checkpoints (see Section \ref{sec:language_model_pret}). For the monolingual experiments we also include a FastText (FT) baseline \cite{joulin-etal-2017-bag}, which relies on monolingual FT embeddings trained on Common Crawl and Wikipedia \cite{grave2018learning} as initialization for each language lookup table.
%In order to finetune the LM to the sentiment analysis task in each language, we adopt the adopter technique \cite{houlsby2019parameter}. Instead of training the whole system, we train only additional layers on top of the language model that is kept frozen during training. We use the adapter configuration proposed in \newcite{pfeiffer2020adapterfusion}. 
%We perform three runs with different seeds for each experiment and we use early stopping on the validation loss. The only parameter search that we perform is on the learning rate (0.001 and 0.0001). 
%.All results are reported according to Macro-averaged F1, averaged on the three run \footnote{Standard deviation and best run results are reported in the supplementary material}.

%\subsection{Comparison systems}

\subsection{Monolingual Evaluation (TweetEval)}
\label{mono:eval}

%\red{Explain evaluation in TweetEval, table with results, etc.}
%As a sanity check of our 
In order to provide an additional point of comparison for our released multilingual language model, we perform an evaluation on standard Twitter-specific tasks in English, for which we can compare its performance with existing models. In particular, we evaluate XLM-Twitter on a suite of seven heterogeneous tweet classification tasks from the TweetEval benchmark \cite{barbieri-etal-2020-tweeteval}. TweetEval is composed of seven tasks: emoji prediction \cite{barbieri2018semeval}, emotion recognition \cite{mohammad2018semeval}, hate speech detection \cite{basile-etal-2019-semeval}, irony detection \cite{van2018semeval}, offensive language identification \cite{zampieri-etal-2019-semeval}, sentiment analysis \cite{rosenthal2019semeval} and stance detection\footnote{The stance detection dataset is in turn split into five subtopics.} \cite{mohammad2016semeval}.

Table \ref{table-results} shows the results of the language models and TweetEval baselines\footnote{Please refer to the original TweetEval paper for details on the implementation of all the baselines.} As can be observed, our proposed XLM-R-Twitter improves over strong baselines such as RoBERTa-base and XLM-R that do not make use of Twitter corpora, and RoBERTa-Twitter, which is trained on Twitter corpora only. This highlights the reliability of our multilingual model in language-specific settings. However, it underperforms when compared with monolingual Twitter-specific models, such as the RoBERTa model futher pre-trained on English tweets proposed in \cite{barbieri-etal-2020-tweeteval}, as well as BERTweet \cite{nguyen-etal-2020-bertweet}, which was trained on a corpus that is an order of magnitude larger.\footnote{While XLM-R-Twitter was fine-tuned on the same amount of English tweets (60M) than RoBERTa-Tw, BERTweet was trained on 850M English tweets.} This is to be expected as goes in line with previous research that shows that multilingual models tend to underperform monolingual models in language-specific tasks \cite{rust2020good}.\footnote{It has been shown that this performance difference could be further decreased by using language-specific tokenizers \cite{rust2020good}, but this was out of scope for this paper.}   In the following section we evaluate XLM-Twitter on multilingual settings, including evaluation in monolingual and cross-lingual scenarios.
%Moreover, while XLM-R-Twitter was fine-tuned on the same amount of English tweets (60M) than RoBERTa-Tw, BERTweet was trained on 850M English tweets.

%plus additional multilingual tweets (138M). While

\subsection{Multilingual Evaluation (Sentiment Analysis)}
\label{cross:eval}

We focus our evaluation on multilingual Sentiment Analysis (SA). %To this end, we first compile and standardize sentiment analysis datasets in different languages, which we explain in the following subsection.
We first flesh out the process followed to compile and unify our cross-lingual SA benchmark (Section \ref{datasets}). %, and then provide architectural details and present the three baselines we evaluate (Section \ref{sec:exp_setting}). 
Our experiments\footnote{Standard deviation and best run results are provided, for completeness, in the appendix.} can then be grouped into two types: when no training in the target language is available, i.e., zero-shot (Section \ref{sec:zeroshot}), and when the evaluated models have access to target language training data, either alone or as part of a larger fully multilingual training set (Section \ref{sec:withtarget}).

%\red{\paragraph{Evaluation metrics.} We XXXXXXXXXXXXXXX TODO: Explain metric}

\begin{table*}[!t]
\centering
\setlength{\tabcolsep}{3.5pt}
\scalebox{0.85}{ 
\begin{tabular}{ccccccccccllllllllc}
\toprule
\textbf{} & \multicolumn{9}{c}{\textbf{XLM-R}} & \multicolumn{9}{c}{\textbf{XLM-Twitter}} \\
\cmidrule(lr){2-10} \cmidrule(lr){11-19}
\multicolumn{1}{l}{\textbf{}} & \textbf{Ar} & \textbf{En} & \textbf{Fr} & \textbf{De} & \textbf{Hi} & \textbf{It} & \textbf{Pt} & \textbf{Es} & \textit{\textbf{All-1}} & \multicolumn{1}{c}{\textbf{Ar}} & \multicolumn{1}{c}{\textbf{En}} & \multicolumn{1}{c}{\textbf{Fr}} & \multicolumn{1}{c}{\textbf{De}} & \multicolumn{1}{c}{\textbf{Hi}} & \multicolumn{1}{c}{\textbf{It}} & \multicolumn{1}{c}{\textbf{Pt}} & \multicolumn{1}{c}{\textbf{Es}} & \textit{\textbf{All-1}} \\
\cmidrule(lr){2-9} \cmidrule(lr){10-10}\cmidrule(lr){11-18} \cmidrule(lr){19-19}

\textbf{Ar} & {\color[HTML]{9B9B9B} 63.6} & \textbf{64.1} & 54.4 & 53.9 & 22.9 & 57.4 & 62.4 & 62.2 & \textit{59.2} & {\color[HTML]{9B9B9B} 67.7} & \textbf{66.6} & 62.1 & 59.3 & 46.3 & 63.0 & 60.1 & 65.3 & \textit{64.3} \\
\textbf{En} & 64.2 & {\color[HTML]{9B9B9B} 68.2} & 61.6 & 63.5 & 23.7 & \textbf{68.1} & 65.9 & 67.8 & \textit{68.2} & 64.0 & {\color[HTML]{9B9B9B} 66.9} & 60.6 & 67.8 & 35.2 & 67.7 & 61.6 & \textbf{68.7} & \textit{70.3} \\
\textbf{Fr} & 45.4 & 52.1 & {\color[HTML]{9B9B9B} 72.0} & 36.5 & 16.7 & 43.3 & 40.8 & \textbf{56.7} & \textit{53.6} & 47.7 & \textbf{59.2} & {\color[HTML]{9B9B9B} 68.2} & 38.7 & 20.9 & 45.1 & 38.6 & 52.5 & \textit{50.0} \\
\textbf{De} & 43.5 & \textbf{64.4} & 55.2 & {\color[HTML]{9B9B9B} 73.6} & 21.5 & 60.8 & 60.1 & 62.0 & \textit{63.6} & 46.5 & 65.0 & 56.4 & {\color[HTML]{9B9B9B} 76.1} & 36.9 & \textbf{66.3} & 65.1 & 65.8 & \textit{65.9} \\
\textbf{Hi} & 48.2 & 52.7 & 43.6 & 47.6 & {\color[HTML]{9B9B9B} 36.6} & \textbf{54.4} & 51.6 & 51.7 & \textit{49.9} & 50.0 & 55.5 & 51.5 & 44.4 & {\color[HTML]{9B9B9B} 40.3} & \textbf{56.1} & 51.2 & 49.5 & \textit{57.8} \\
\textbf{It} & 48.8 & 65.7 & 63.9 & \textbf{66.9} & 22.1 & {\color[HTML]{9B9B9B} 71.5} & 63.1 & 58.9 & \textit{65.7} & 41.9 & 59.6 & 60.8 & 64.5 & 24.6 & {\color[HTML]{9B9B9B} 70.9} & \textbf{64.7} & 55.1 & \textit{65.2} \\
\textbf{Pt} & 41.5 & 63.2 & 57.9 & 59.7 & 26.5 & 59.6 & {\color[HTML]{9B9B9B} 67.1} & \textbf{65.0} & \textit{65.0} & 56.4 & \textbf{67.7} & 62.8 & 64.4 & 26.0 & 67.1 & {\color[HTML]{9B9B9B} 76.0} & 64.0 & \textit{71.4} \\
\textbf{Es} & 47.1 & 63.1 & 56.8 & 57.2 & 26.2 & 57.6 & \textbf{63.1} & {\color[HTML]{9B9B9B} 65.9} & \textit{63.0} & 52.9 & 66.0 & 64.5 & 58.7 & 30.7 & 62.4 & \textbf{67.9} & {\color[HTML]{9B9B9B} 68.5} & \textit{66.2}

\\
\bottomrule
\end{tabular}
}
\caption{\label{tab:cross-zero} Zero-shot cross-lingual sentiment analysis results (F1). We use the best model in the language on the column and evaluate on the test set of the language of each row. For example, when we forward the best XLM-R trained on English text on the Arabic test set we obtain 64.1. In the columns \textit{All minus one (\textbf{All-1})} we train on all the languages excluding the one of each row. For example, we obtain a F1 of 59.2 on the Arabic test set when we train an XLM-R using all the languages excluding Arabic. On the diagonals, in gray, models are trained and evaluated on the same language.}
\end{table*}

\begin{table}[]
\centering
\setlength{\tabcolsep}{2.2pt}
\resizebox{\columnwidth}{!}{
%\scalebox{0.90}{ 
\begin{tabular}{llcc}
\toprule
\multicolumn{1}{c}{\textbf{Lang.}} & \multicolumn{1}{c}{\textbf{Dataset}} & \textbf{Time-Train} & \textbf{Time-Test} \\
\cmidrule(lr){1-4}
Arabic & SemEval-17 \cite{rosenthal2017semeval} & 09/16-11/16 & 12/16-1/17 \\
English & SemEval-17 \cite{rosenthal2017semeval} & 01/12-12/15 & 12/16-1/17 \\
French & Deft-17 \cite{benamara2017analyse} & 2014-2016 & Same \\
German & SB-10K \cite{cieliebak2017twitter} & 8/13-10/13 & Same \\
Hindi & SAIL 2015 \cite{patra2015shared} & NA,3-month & Same \\
Italian & Sentipolc-16 \cite{barbieri2016overview} & 2013-2016 & 2016 \\
Portug. & SentiBR \cite{brum2017building} & 1/17-7/17 & Same \\
Spanish & Intertass %2017 
\cite{diaz2018democratization} & 7/16-01/17 & Same \\ 
\bottomrule
\end{tabular}
}
\caption{\label{tab:datasets} Sentiment analysis datasets for the eight languages used in our experiments.}
\end{table}

%as initialization of the lookup table in each language the 

\subsubsection{Unified Multilingual Sentiment Analysis Benchmark (UMSAB)}
\label{datasets}

\vspace{1ex plus 0.5ex}

%\paragraph{Dataset} 
We aim at constructing a balanced multilingual SA dataset, i.e., where all languages are equally distributed in terms of frequency, and with representation of typologically distant languages. To this end, we compiled monolingual SA datasets for eight diverse languages. We list the languages and relevant statistics in Table \ref{tab:datasets}, as well as their spanning timeframes. Given that retaining the original distribution would skew the unified dataset towards the most frequent languages, we established a maximum number of tweets corresponding to the size of the smallest dataset, specifically the 3,033 for the Hindi portion, and prune all data splits for all languages with this threshold. This leaves 1,839 training tweets (with 15\% of them allocated to a fixed validation set), and 870 for testing. The total size of the dataset is thus 24,262 tweets. Let us highlight two additional important design decisions: first, we enforced a balanced distribution across the three labels (positive, negative and neutral), and second, we kept the original training/test splits in each dataset. After this preprocessing, we obtain 8 datasets of 3,033 instances, respectively. Note that some languages in this dataset agglutinate or refer to specific variations. In particular, we use Hindi to refer to the grouping of Hindi, Bengalu and Tamil, Portuguese for Brazilian Portuguese, and Spanish for Iberian, Peruvian and Costa Rican variations. 

% \begin{figure}[!t]
% \centering
% \includegraphics[width=\columnwidth,height=6cm]{acl-ijcnlp2021-templates/imgs/xlm-twitter-sims2.png}
% \caption{Cross-lingual similarity (by cosine distance) between the English training set and the test sets in the other 7 languages. The embeddings are obtained by averaging all the transformer layers for each token.}
% \label{fig:vis}
% \end{figure}

%\red{Hindi= Hindi, Bengalu, Tamil (55.67 \%, 43.2 \%, 39.28 \%)}
%\red{Portuguese is Brasilian Portuguese}
%\red{Spanish is collection of Ibierian, Peruvian and Costa Rica Spanish}

%In order to standardize the datasets and perform cross-lingual controlled experiments, we decided to (1) keep the distribution of the three labels (positive, negative and neutral) balanced, (2) keep the same number of tweets per dataset. (3) keep the same original training/test splits of each dataset %\red{EXPLAIN BETTER: (1) to avoid to learn distribution instead of language features, (2) to avoid unfairness when a model is trained with more data than another}

%Explain Train in general domain and then twitter, cite our Findings paper to justify (+others), etc.

%\section{Sentiment Analysis}

%\subsection{Datasets}

%\subsection{Data}

%\paragraph{Sentiment analysis datasets}

%\paragraph{Unification and statistics}

\subsubsection{Zero-shot Cross-lingual Transfer}
\label{sec:zeroshot}

 %\citep{rasooli-tetrault-2015}. 
 
%We report results in two main settings, one in which the target language is considered (Section \ref{sec:zeroshot}) and another one in a purely zero-shot cross-lingual setting (Section \ref{sec:withtarget}).

%\subsubsection{Zero-shot Cross-lingual Transfer}

%In this setting we investigate the capabilities of XLM-Twitter in zero-shot cross-lingual transfer (i.e., train on a language and evaluate on different language). 
%\luis{cita a lample etc al no ser LMs no la veo mucho}
%Multilingual language models have already shown their potential for many cross-lingual tasks \cite{lample2019cross,pires-etal-2019-multilingual,artetxe2020cross}, and in this experiment we specifically test their potential in cross-lingual sentiment analysis. 
Table \ref{tab:cross-zero} shows zero-shot results of XLM-R and XLM-Twitter in our multilingual sentiment analysis benchmark. The performance of both models is competitive, especially considering the diversity of domains\footnote{For instance, for Arabic we find trending topics such as iPhone or vegetarianism, where the Portuguese dataset is dominated by comments on TV shows.} and that the source language was not seen during training. An interesting observation concerns those cases in which zero-shot models outperform their monolingual counterparts (e.g., English$	\rightarrow$Arabic or Italian$\rightarrow$Hindi). Additionally, XLM-Twitter proves more robust, achieving the best overall results in six of the eight languages, with consistent improvements in general, and with remarkable improvements in e.g., Hindi, outperforming XLM-R by 7.9 absolute points. Finally, let us provide some insights on the results obtained in an all-minus-one (the \textbf{All-1} columns in Table \ref{tab:cross-zero}) setting. Here, notable cases are, first, Hindi, in which XLM-R and XLM-Twitter models benefit substantially by having access to more training data, with this improvement being more pronounced in XLM-Twitter. Second, the results for the English dataset suggest that compiling a larger training set helps, although this may be also attributed to identical tokens shared between English and the other languages, such as named entities, hashtags or colloquialisms and slang.

\subsubsection{Cross-lingual Transfer with Target Language Training Data}
\label{sec:withtarget}

\vspace{1ex plus 0.5ex}

Table \ref{tab:expwithtarget} shows macro-F1 results for the following three settings: (1) \textbf{monolingual}, where we train and test in one single language; (2) \textbf{bilingual}, where we use the best-performing cross-lingual zero-shot model, and continue fine-tuning on training data from the target language; and (3) an entirely \textbf{multilingual} setting where we train with data from all languages. One of the most notable conclusions in the light of these figures is that increasing the training data even in different languages is a useful strategy, and is particularly rewarding in the case of XLM-Twitter and in challenging datasets and languages (e.g., the Hindi results significantly increase from 40.29 to 56.39). Interestingly, a smart selection of languages based on validation accuracy achieves better results than if trained on all languages in half of the cases. This may be due to the (dis)similarity of the datasets (in terms of topic or typological proximity), although overall the main conclusion we can draw is that there is an obvious trade-off, as a single multilingual model is often more practical and versatile. %As an additional qualitative analysis, we plot in Figure \ref{fig:vis} a sample of similarity scores (by cosine distance) between XLM-Twitter-based embeddings obtained from the English \textit{training set} and the test sets for the other 7 languages. In addition to the clearly low resemblance with Hindi, we find that the most similar languages in the embedding space are English are French, suggesting that not only typology, but also topic overlap, may play an important role in the quality of these multilingual representations.

\begin{table}[!t]
\centering
\setlength{\tabcolsep}{2.2pt}
\resizebox{\columnwidth}{!}{
%\scalebox{0.91}{ 
\begin{tabular}{cccccccc}
\toprule
 & \multicolumn{3}{c}{Monolingual} & \multicolumn{2}{c}{Bilingual} & \multicolumn{2}{c}{Multilingual} \\
 \cmidrule(lr){2-4} \cmidrule(lr){5-6} \cmidrule(lr){7-8}
\textbf{} & \textbf{FT} & \multicolumn{1}{c}{\textbf{XLM-R}} & \multicolumn{1}{c}{\textbf{XLM-T}} & \multicolumn{1}{c}{\textbf{XLM-R}} & \multicolumn{1}{c}{\textbf{XLM-T}} & \multicolumn{1}{c}{\textbf{XLM-R}} & \multicolumn{1}{c}{\textbf{XLM-T}} \\
\cmidrule(lr){2-4} \cmidrule(lr){5-6} \cmidrule(lr){7-8}

\textbf{Ar} & 45.98 & 63.56 & \textbf{67.67} & 63.63 {\small (En)} & 67.65 {\small (En)} & 64.31 & 66.89 \\
\textbf{En} & 50.85 & 68.18 & 66.89 & 65.07 {\small (It)} & 67.47 {\small (Es)} & 68.52 & \textbf{70.63} \\
\textbf{Fr} & 54.82 & 71.98 & 68.19 & \textbf{73.55} {\small (Sp)} & 68.24 {\small (En)} & 70.52 & 71.18 \\
\textbf{De} & 59.56 & 73.61 & 76.13 & 72.48 {\small (En)} & 75.49 {\small (It)} & 72.84 & \textbf{77.35} \\
\textbf{Hi} & 37.08 & 36.60 & 40.29 & 33.57 {\small (It)} & 55.35 {\small (It)} & 53.39 & \textbf{56.39} \\
\textbf{It} & \multicolumn{1}{l}{54.65} & 71.47 & 70.91 & 70.43 {\small (Ge)} & \textbf{73.50} {\small (Pt)} & 68.62 & 69.06 \\
\textbf{Pt} & \multicolumn{1}{l}{55.05} & 67.11 & 75.98 & 71.87 {\small (Sp)} & \textbf{76.08} {\small (En)} & 69.79 & 75.42 \\
\textbf{Sp} & \multicolumn{1}{l}{50.06} & 65.87 & 68.52 & 67.68 {\small (Po)} & \textbf{68.68} {\small (Pt)} & 66.03 & 67.91 

\\
\cmidrule(lr){2-4} \cmidrule(lr){5-6} \cmidrule(lr){7-8}
%\cmidrule(lr){1-8}
\textbf{All} & \multicolumn{1}{l}{51.01} & 64.80 & 66.82 & 64.78 & 69.06 & 66.75 & \textbf{69.35} \\
\bottomrule
\end{tabular}
}
%}
\caption{\label{tab:expwithtarget} Cross-lingual sentiment analysis F1 results on target languages using target language training data (Monolingual) only, combined with training data from another language (Bilingual) and with all languages at once (Multilingual). "All" is computed as the average of all individual results.
%\red{TODO: Add bold numbers, remove one decimal and make it one column. Esta abajo commented esta tabla, pero se vee muy pequena.}
}
\end{table}
 
\subsection{Qualitative Analysis}
\label{qualitative}

\vspace{1ex plus 0.5ex}

\begin{figure}[!t]
\hspace{-0.33cm}
\scalebox{0.85}{ 
\begin{tikzpicture}
\begin{axis}[ 
    width=9.4cm,
    height=7.5cm,
    ylabel shift = -0.06cm,
    ylabel near ticks,    
    xlabel={Similarity},
    ylabel={Density},
    xmin=0.88, xmax=0.99,
    ymin=0, ymax=60,
    xtick={0.8, 0.825, 0.85, 0.875, 0.9, 0.925, 0.95, 0.975, 1},
    ytick={0,10,20,30,40,50},
    legend pos=north west,
    ymajorgrids=true,
    grid style=dashed,
    %line width=5pt,
    %every picture/.style={thick} %or use: "`line width=1 pt,"<-- note:if you write line width, you must use a value with unit
]

\definecolor{c1}{RGB}{145, 26, 213}
\definecolor{c2}{RGB}{23,88,202}
\definecolor{c3}{RGB}{80, 189, 230}
\definecolor{c4}{RGB}{90,200,101}
\definecolor{c5}{RGB}{249,231,33}
\definecolor{c6}{RGB}{230, 126, 34}
\definecolor{c7}{RGB}{176, 58, 46}

\newcommand{\mywidth}{1.2pt}
%AR
\addplot[color=c1, line width=\mywidth]
    coordinates {
(0.8676,0.0018) (0.8686,0.006) (0.8697,0.0162) (0.8707,0.0368) (0.8717,0.07) (0.8728,0.1112) (0.8738,0.1476) (0.8748,0.1638) (0.8758,0.152) (0.8769,0.1179) (0.8779,0.0764) (0.8789,0.0414) (0.88,0.0188) (0.881,0.0071) (0.882,0.0024) (0.8831,0.0012) (0.8841,0.0024) (0.8851,0.0076) (0.8861,0.0213) (0.8872,0.0505) (0.8882,0.1013) (0.8892,0.1719) (0.8903,0.2472) (0.8913,0.3015) (0.8923,0.3121) (0.8934,0.2742) (0.8944,0.2043) (0.8954,0.1291) (0.8965,0.0691) (0.8975,0.0314) (0.8985,0.0128) (0.8995,0.007) (0.9006,0.0109) (0.9016,0.0265) (0.9026,0.0607) (0.9037,0.1231) (0.9047,0.2252) (0.9057,0.3796) (0.9068,0.5977) (0.9078,0.8811) (0.9088,1.2106) (0.9098,1.5466) (0.9109,1.8468) (0.9119,2.0888) (0.9129,2.2785) (0.914,2.4405) (0.915,2.6078) (0.916,2.8205) (0.9171,3.1326) (0.9181,3.6143) (0.9191,4.3337) (0.9201,5.3208) (0.9212,6.5378) (0.9222,7.8919) (0.9232,9.2901) (0.9243,10.7079) (0.9253,12.224) (0.9263,13.9862) (0.9274,16.0963) (0.9284,18.4797) (0.9294,20.8636) (0.9305,22.9265) (0.9315,24.51) (0.9325,25.7008) (0.9335,26.7353) (0.9346,27.8568) (0.9356,29.2513) (0.9366,31.0418) (0.9377,33.2667) (0.9387,35.8398) (0.9397,38.5525) (0.9408,41.141) (0.9418,43.3666) (0.9428,45.0412) (0.9438,45.9902) (0.9449,46.0254) (0.9459,44.9901) (0.9469,42.8372) (0.948,39.6479) (0.949,35.5878) (0.95,30.8843) (0.9511,25.8497) (0.9521,20.8671) (0.9531,16.2945) (0.9541,12.3579) (0.9552,9.1294) (0.9562,6.5796) (0.9572,4.6345) (0.9583,3.2006) (0.9593,2.1752) (0.9603,1.4569) (0.9614,0.9581) (0.9624,0.6116) (0.9634,0.3721) (0.9644,0.2109) (0.9655,0.1086) (0.9665,0.0497) (0.9675,0.0199) (0.9686,0.0069) (0.9696,0.002)
    };

%FR
\addplot[color=c2, line width=\mywidth]
    coordinates {
(0.8471,0.0022) (0.8483,0.0119) (0.8496,0.0432) (0.8508,0.1061) (0.8521,0.1761) (0.8533,0.1978) (0.8546,0.1502) (0.8558,0.0772) (0.8571,0.0269) (0.8583,0.0063) (0.8596,0.001) (0.8608,0.0001) (0.8621,0.0) (0.8634,0.0) (0.8646,0.0) (0.8659,0.0) (0.8671,0.0) (0.8684,0.0) (0.8696,0.0) (0.8709,0.0) (0.8721,0.0) (0.8734,0.0) (0.8746,0.0) (0.8759,0.0) (0.8771,0.0) (0.8784,0.0) (0.8796,0.0) (0.8809,0.0) (0.8821,0.0) (0.8834,0.0) (0.8846,0.0) (0.8859,0.0) (0.8871,0.0) (0.8884,0.0) (0.8896,0.0) (0.8909,0.0) (0.8921,0.0) (0.8934,0.0) (0.8946,0.0) (0.8959,0.0) (0.8971,0.0) (0.8984,0.0) (0.8996,0.0) (0.9009,0.0001) (0.9021,0.001) (0.9034,0.0063) (0.9046,0.0274) (0.9059,0.0825) (0.9071,0.1754) (0.9084,0.273) (0.9096,0.3257) (0.9109,0.3091) (0.9121,0.2435) (0.9134,0.1885) (0.9146,0.1927) (0.9159,0.2406) (0.9171,0.3012) (0.9184,0.4083) (0.9196,0.6162) (0.9209,0.9102) (0.9221,1.2095) (0.9234,1.4303) (0.9246,1.5989) (0.9259,1.9689) (0.9271,2.8543) (0.9284,4.2472) (0.9296,5.8609) (0.9309,7.5761) (0.9322,9.5235) (0.9334,11.7726) (0.9347,14.2549) (0.9359,17.0075) (0.9372,20.3767) (0.9384,24.9794) (0.9397,31.2388) (0.9409,38.6633) (0.9422,45.8231) (0.9434,51.1365) (0.9447,53.5986) (0.9459,53.4552) (0.9472,52.2048) (0.9484,51.0932) (0.9497,49.9626) (0.9509,47.8097) (0.9522,43.8886) (0.9534,38.3517) (0.9547,31.9823) (0.9559,25.462) (0.9572,19.1912) (0.9584,13.5625) (0.9597,8.9809) (0.9609,5.6314) (0.9622,3.4164) (0.9634,2.049) (0.9647,1.1867) (0.9659,0.6093) (0.9672,0.2525) (0.9684,0.0793) (0.9697,0.0182) (0.9709,0.003)
    };
%DE
\addplot[color=c3, line width=\mywidth]
    coordinates {
(0.7933,0.001) (0.7953,0.0031) (0.7972,0.0083) (0.7991,0.0188) (0.8011,0.0356) (0.803,0.0569) (0.805,0.0762) (0.8069,0.086) (0.8089,0.0822) (0.8108,0.0688) (0.8128,0.0559) (0.8147,0.055) (0.8166,0.0732) (0.8186,0.1083) (0.8205,0.1492) (0.8225,0.1803) (0.8244,0.1912) (0.8264,0.1831) (0.8283,0.1655) (0.8302,0.1485) (0.8322,0.1365) (0.8341,0.1304) (0.8361,0.1305) (0.838,0.1384) (0.84,0.1562) (0.8419,0.1863) (0.8439,0.2283) (0.8458,0.2757) (0.8477,0.3176) (0.8497,0.3489) (0.8516,0.3773) (0.8536,0.419) (0.8555,0.4828) (0.8575,0.5582) (0.8594,0.6203) (0.8614,0.6473) (0.8633,0.635) (0.8652,0.5968) (0.8672,0.5495) (0.8691,0.502) (0.8711,0.4575) (0.873,0.4225) (0.875,0.4072) (0.8769,0.415) (0.8788,0.4342) (0.8808,0.4461) (0.8827,0.441) (0.8847,0.4295) (0.8866,0.4358) (0.8886,0.4796) (0.8905,0.5607) (0.8925,0.6582) (0.8944,0.7462) (0.8963,0.8129) (0.8983,0.8691) (0.9002,0.9413) (0.9022,1.0562) (0.9041,1.2271) (0.9061,1.4488) (0.908,1.7038) (0.91,1.9765) (0.9119,2.2692) (0.9138,2.6082) (0.9158,3.0376) (0.9177,3.6012) (0.9197,4.3208) (0.9216,5.1879) (0.9236,6.182) (0.9255,7.3103) (0.9275,8.6365) (0.9294,10.2684) (0.9313,12.3061) (0.9333,14.796) (0.9352,17.7231) (0.9372,21.0271) (0.9391,24.5937) (0.9411,28.2177) (0.943,31.5857) (0.9449,34.312) (0.9469,36.0027) (0.9488,36.3224) (0.9508,35.068) (0.9527,32.2521) (0.9547,28.1528) (0.9566,23.278) (0.9586,18.236) (0.9605,13.5729) (0.9624,9.6518) (0.9644,6.6139) (0.9663,4.4131) (0.9683,2.8912) (0.9702,1.8598) (0.9722,1.1601) (0.9741,0.6858) (0.9761,0.3741) (0.978,0.1838) (0.9799,0.0798) (0.9819,0.0302) (0.9838,0.0099) (0.9858,0.0028)
    };
%Hi
\addplot[color=c4, line width=\mywidth]
    coordinates {
(0.5197,0.0004) (0.5247,0.0011) (0.5297,0.0028) (0.5347,0.0063) (0.5397,0.0126) (0.5447,0.0222) (0.5497,0.0345) (0.5547,0.048) (0.5597,0.0601) (0.5647,0.0689) (0.5697,0.0737) (0.5747,0.0754) (0.5797,0.0753) (0.5847,0.0745) (0.5897,0.0732) (0.5947,0.0714) (0.5998,0.0694) (0.6048,0.0681) (0.6098,0.0685) (0.6148,0.0709) (0.6198,0.0747) (0.6248,0.078) (0.6298,0.0789) (0.6348,0.0761) (0.6398,0.0699) (0.6448,0.0618) (0.6498,0.0534) (0.6548,0.0458) (0.6598,0.0406) (0.6648,0.0399) (0.6698,0.0458) (0.6748,0.0592) (0.6798,0.0781) (0.6848,0.0987) (0.6898,0.1175) (0.6948,0.1343) (0.6998,0.1522) (0.7048,0.1757) (0.7098,0.207) (0.7148,0.2442) (0.7198,0.2816) (0.7248,0.3131) (0.7298,0.3351) (0.7348,0.348) (0.7399,0.3549) (0.7449,0.3583) (0.7499,0.3585) (0.7549,0.3524) (0.7599,0.3371) (0.7649,0.3131) (0.7699,0.2865) (0.7749,0.2671) (0.7799,0.2648) (0.7849,0.2843) (0.7899,0.3234) (0.7949,0.3743) (0.7999,0.4278) (0.8049,0.4777) (0.8099,0.5222) (0.8149,0.5627) (0.8199,0.6001) (0.8249,0.6328) (0.8299,0.6584) (0.8349,0.6772) (0.8399,0.6945) (0.8449,0.7208) (0.8499,0.769) (0.8549,0.8489) (0.8599,0.9641) (0.8649,1.1144) (0.8699,1.3041) (0.8749,1.5482) (0.88,1.8721) (0.885,2.3037) (0.89,2.8658) (0.895,3.5741) (0.9,4.4402) (0.905,5.4741) (0.91,6.6808) (0.915,8.0479) (0.92,9.5261) (0.925,11.014) (0.93,12.3601) (0.935,13.3953) (0.94,13.9845) (0.945,14.0661) (0.95,13.6497) (0.955,12.7737) (0.96,11.4701) (0.965,9.7792) (0.97,7.8053) (0.975,5.7468) (0.98,3.8526) (0.985,2.3276) (0.99,1.2577) (0.995,0.6045) (1.0,0.2575) (1.005,0.097) (1.01,0.0322) (1.015,0.0094)
    };
%IT
\addplot[color=c5, line width=\mywidth]
    coordinates {
(0.8319,0.0013) (0.8334,0.0042) (0.8348,0.0114) (0.8362,0.0257) (0.8377,0.0489) (0.8391,0.078) (0.8405,0.1044) (0.842,0.1178) (0.8434,0.114) (0.8448,0.0998) (0.8463,0.091) (0.8477,0.1028) (0.8491,0.1398) (0.8506,0.191) (0.852,0.2349) (0.8534,0.254) (0.8549,0.2483) (0.8563,0.2354) (0.8578,0.2358) (0.8592,0.2571) (0.8606,0.2891) (0.8621,0.3112) (0.8635,0.3067) (0.8649,0.275) (0.8664,0.2349) (0.8678,0.2155) (0.8692,0.2393) (0.8707,0.3085) (0.8721,0.404) (0.8735,0.4961) (0.875,0.5585) (0.8764,0.5788) (0.8778,0.5629) (0.8793,0.5343) (0.8807,0.5254) (0.8821,0.5591) (0.8836,0.634) (0.885,0.7273) (0.8865,0.813) (0.8879,0.8773) (0.8893,0.918) (0.8908,0.9334) (0.8922,0.9166) (0.8936,0.8618) (0.8951,0.7751) (0.8965,0.6768) (0.8979,0.5918) (0.8994,0.5362) (0.9008,0.5183) (0.9022,0.5517) (0.9037,0.6642) (0.9051,0.8847) (0.9065,1.2192) (0.908,1.64) (0.9094,2.1034) (0.9109,2.5822) (0.9123,3.0858) (0.9137,3.6503) (0.9152,4.3124) (0.9166,5.0867) (0.918,5.9551) (0.9195,6.8706) (0.9209,7.7889) (0.9223,8.7281) (0.9238,9.8087) (0.9252,11.2186) (0.9266,13.1208) (0.9281,15.5796) (0.9295,18.5515) (0.9309,21.915) (0.9324,25.5097) (0.9338,29.1839) (0.9352,32.84) (0.9367,36.4201) (0.9381,39.797) (0.9396,42.6582) (0.941,44.5137) (0.9424,44.857) (0.9439,43.3704) (0.9453,40.0571) (0.9467,35.2607) (0.9482,29.5834) (0.9496,23.7265) (0.951,18.3026) (0.9525,13.6971) (0.9539,10.0347) (0.9553,7.2451) (0.9568,5.169) (0.9582,3.6412) (0.9596,2.5244) (0.9611,1.7116) (0.9625,1.1223) (0.9639,0.7003) (0.9654,0.4081) (0.9668,0.2183) (0.9683,0.1054) (0.9697,0.0454) (0.9711,0.0172) (0.9726,0.0057) (0.974,0.0016)
    };
%PT
\addplot[color=c6, line width=\mywidth]
    coordinates {
(0.8666,0.0015) (0.8677,0.0041) (0.8687,0.0099) (0.8698,0.0213) (0.8709,0.0401) (0.872,0.0671) (0.873,0.1004) (0.8741,0.1355) (0.8752,0.1682) (0.8763,0.1969) (0.8773,0.2229) (0.8784,0.2476) (0.8795,0.2687) (0.8805,0.2792) (0.8816,0.2711) (0.8827,0.2412) (0.8838,0.1942) (0.8848,0.1406) (0.8859,0.0919) (0.887,0.0567) (0.888,0.039) (0.8891,0.0406) (0.8902,0.0632) (0.8913,0.1088) (0.8923,0.179) (0.8934,0.2723) (0.8945,0.3837) (0.8956,0.505) (0.8966,0.6258) (0.8977,0.7352) (0.8988,0.8241) (0.8998,0.8896) (0.9009,0.9396) (0.902,0.9938) (0.9031,1.0781) (0.9041,1.2144) (0.9052,1.4121) (0.9063,1.668) (0.9073,1.9726) (0.9084,2.3165) (0.9095,2.6885) (0.9106,3.067) (0.9116,3.4138) (0.9127,3.6844) (0.9138,3.8515) (0.9149,3.9289) (0.9159,3.9725) (0.917,4.0551) (0.9181,4.2292) (0.9191,4.5029) (0.9202,4.8471) (0.9213,5.227) (0.9224,5.6364) (0.9234,6.108) (0.9245,6.6959) (0.9256,7.4434) (0.9266,8.3616) (0.9277,9.4278) (0.9288,10.6025) (0.9299,11.8505) (0.9309,13.1556) (0.932,14.5238) (0.9331,15.9746) (0.9342,17.5254) (0.9352,19.1774) (0.9363,20.9142) (0.9374,22.7253) (0.9384,24.6408) (0.9395,26.7483) (0.9406,29.1624) (0.9417,31.9554) (0.9427,35.0897) (0.9438,38.392) (0.9449,41.5723) (0.9459,44.2668) (0.947,46.0942) (0.9481,46.7254) (0.9492,45.9662) (0.9502,43.8208) (0.9513,40.4988) (0.9524,36.3493) (0.9535,31.7526) (0.9545,27.0257) (0.9556,22.3872) (0.9567,17.9828) (0.9577,13.9337) (0.9588,10.3628) (0.9599,7.3794) (0.961,5.0417) (0.962,3.3294) (0.9631,2.1487) (0.9642,1.3672) (0.9653,0.857) (0.9663,0.5218) (0.9674,0.3019) (0.9685,0.1623) (0.9695,0.0795) (0.9706,0.0351) (0.9717,0.0138) (0.9728,0.0048)
    };
%ES
\addplot[color=c7, line width=\mywidth]
    coordinates {
(0.8988,0.003) (0.8995,0.0075) (0.9003,0.0172) (0.901,0.0358) (0.9017,0.0675) (0.9024,0.1157) (0.9032,0.1802) (0.9039,0.2561) (0.9046,0.3332) (0.9054,0.3994) (0.9061,0.4453) (0.9068,0.4685) (0.9076,0.4759) (0.9083,0.4825) (0.909,0.5071) (0.9098,0.5678) (0.9105,0.6769) (0.9112,0.8367) (0.9119,1.0383) (0.9127,1.2625) (0.9134,1.4864) (0.9141,1.6912) (0.9149,1.871) (0.9156,2.0364) (0.9163,2.2107) (0.9171,2.4198) (0.9178,2.6798) (0.9185,2.9878) (0.9193,3.3214) (0.92,3.6473) (0.9207,3.9361) (0.9215,4.1756) (0.9222,4.376) (0.9229,4.5643) (0.9236,4.771) (0.9244,5.0199) (0.9251,5.3261) (0.9258,5.7031) (0.9266,6.1693) (0.9273,6.7438) (0.928,7.4322) (0.9288,8.2134) (0.9295,9.0401) (0.9302,9.8601) (0.931,10.6472) (0.9317,11.4216) (0.9324,12.2443) (0.9332,13.1878) (0.9339,14.299) (0.9346,15.5801) (0.9353,17.0016) (0.9361,18.5395) (0.9368,20.2152) (0.9375,22.1075) (0.9383,24.3255) (0.939,26.9514) (0.9397,29.9833) (0.9405,33.3085) (0.9412,36.7213) (0.9419,39.9793) (0.9427,42.8729) (0.9434,45.2819) (0.9441,47.1928) (0.9448,48.6733) (0.9456,49.8157) (0.9463,50.6784) (0.947,51.2505) (0.9478,51.4504) (0.9485,51.1519) (0.9492,50.226) (0.95,48.5863) (0.9507,46.2255) (0.9514,43.2316) (0.9522,39.7734) (0.9529,36.0567) (0.9536,32.2704) (0.9544,28.5496) (0.9551,24.9704) (0.9558,21.5725) (0.9565,18.3877) (0.9573,15.4544) (0.958,12.8137) (0.9587,10.4957) (0.9595,8.5085) (0.9602,6.8361) (0.9609,5.4447) (0.9617,4.2917) (0.9624,3.3348) (0.9631,2.539) (0.9639,1.8795) (0.9646,1.3409) (0.9653,0.9139) (0.9661,0.5901) (0.9668,0.3582) (0.9675,0.203) (0.9682,0.1069) (0.969,0.052) (0.9697,0.0233) (0.9704,0.0096) (0.9712,0.0036)
    };

\legend{Ar, Fr, De, Hi, It, Pt, Es}
    
\end{axis}
\end{tikzpicture}
}
%\vspace{-0.5cm}
\caption{\label{fig:vis} Cross-lingual similarity (by cosine distance) between the English training set and the test sets in the other 7 languages. The embeddings are obtained by averaging all the XLM-Twitter contextualized embeddings for each tweet.}

\end{figure}
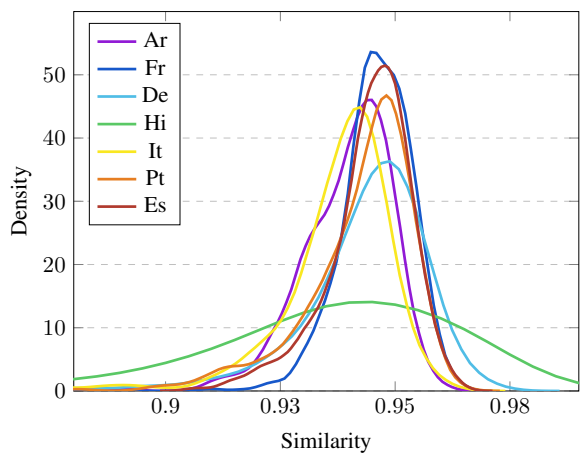

As an additional qualitative analysis, we plot in Figure \ref{fig:vis} a sample of similarity scores (by cosine distance) between XLM-Twitter-based embeddings obtained from the English \textit{training set} and the sentiment analysis test sets for the other 7 languages (see Section \ref{datasets}). In addition to the clearly low resemblance with Hindi, we find that the most similar languages in the embedding space are English and French, suggesting that not only typology, but also topic overlap, may play an important role in the quality of these multilingual representations. This becomes even more apparent in Arabic, which differs from English in typology and script, but has similar representations. The Arabic and English datasets were obtained using the same keywords.%$ for tweet retrieval.

%\subsubsection{Few-shot cross-lingual transfer}
% NO HAY FEW SHOT
%\subsubsection{Multilingual transfer}

\section{Conclusions}

We have presented a comprehensive framework for %framework and an analysis on the cross-lingual capabilities of 
Twitter-based multilingual LMs, including the release of a new multilingual LM trained on almost 200M tweets. As main test bed for our multilingual experiments, we focused on sentiment analysis, for which we collected datasets in eight languages. After a unification and standardization of the evaluation benchmark, we compared the Twitter-based multilingual language model with a standard multilingual language model trained on general-domain corpora. This multilingual language model along with starting and evaluation code are released to facilitate research in Twitter at a multilingual scale (over thirty languages used for training data).

The results highlight the potential of the domain-specific language model, as more suited to handle social media and specifically multilingual SA. Finally, our analysis reveals trends and potential for this Twitter-based multilingual language model in zero-shot cross-lingual settings when language-specific training data is not available. For future work we are planning to extend this analysis to more languages and tasks, but also to deepen the cross-lingual zero and few shot analysis, particularly focusing on typologically similar languages. Finally, and due to the seasonal nature of Twitter, it would also be interesting to explore correlations between topic distribution and trends and performance in downstream applications.

\section*{Acknowledgments}

We would like to thank Eugenio Martínez Cámara for his involvement in the first stages of this project. Jose Camacho-Collados is supported with a UKRI Future Leaders Fellowship.

\section{Bibliographical References}\label{reference}
%\label{main:ref}

\bibliographystyle{lrec2022-bib}
\bibliography{anthology,lrec2022-example}

%\section{Language Resource References}
%\label{lr:ref}
%\bibliographystylelanguageresource{lrec2022-bib}
%\bibliographylanguageresource{languageresource}

\newpage

\appendix

\section{Full Experimental Results}

This appendix includes the full experimental results, including standard deviation after three runs and the best runs according to the validation set. Table \ref{tab:monolingual} includes the monolingual results; Table \ref{tab:cross-lingual}, the cross-lingual results; and Table \ref{tab:multilingual}, the multilingual experiments.

\newpage

%Introduction \cite{barbieri-etal-2020-tweeteval}

\begin{table}[!t]
\scalebox{0.9}{ 
\begin{tabular}{rcccc}
\toprule
\multicolumn{1}{c}{} & \multicolumn{2}{c}{\textbf{XLM}} & \multicolumn{2}{c}{\textbf{XLM-Twitter}} \\
\cmidrule(lr){2-3} \cmidrule(lr){4-5}
\multicolumn{1}{c}{} & \textbf{F1 macro} & \multicolumn{1}{c}{\textbf{F1 Best}} & \textbf{F1 macro} & \multicolumn{1}{c}{\textbf{F1 Best}} \\
%\cmidrule(){1-5}
\cmidrule(lr){2-3} \cmidrule(lr){4-5}
\textbf{Ar} & 63.56 {\small $\pm$1.29} & 64.89 & \textbf{67.67 {\small $\pm$1.25}} & 69.03 \\
\textbf{En} & \textbf{68.18 {\small $\pm$2.57}} & 69.64 & 66.89 {\small $\pm$1.19} & 67.82 \\
\textbf{Fr} & \textbf{71.98 {\small $\pm$1.46}} & 72.86 & 68.19 {\small $\pm$1.55} & 69.20 \\
\textbf{De} & 73.61 {\small $\pm$0.22} & 73.75 & \textbf{76.13 {\small $\pm$0.53}} & 76.58 \\
\textbf{Hi} & 36.6 {\small $\pm$4.36} & 41.46 & \textbf{40.29 {\small $\pm$7.37}} & 48.79 \\
\textbf{It} & \textbf{71.47 {\small $\pm$1.35}} & 73.02 & 70.91 {\small $\pm$0.87} & 71.41 \\
\textbf{Pt} & 67.11 {\small $\pm$1.1} & 67.89 & \textbf{75.98 {\small $\pm$0.03}} & 76.01 \\
\textbf{Es} & 65.87 {\small $\pm$1.67} & 67.75 & \textbf{68.52 {\small $\pm$0.69}} & 69.01 \\
\bottomrule
\end{tabular}
}
\caption{\label{tab:monolingual} Monolingual experiments. XLM and XLM-Twitter are finetuned for each language. F1 macro is the average of three runs and F1 best is the best one of them.}
\end{table}

\begin{table}[!t]
\setlength{\tabcolsep}{2.5pt}
\scalebox{0.85}{ 
\begin{tabular}{ccccccc}
\toprule
 & \multicolumn{3}{c}{\textbf{XLM}} & \multicolumn{3}{c}{\textbf{XLM-Twitter}} \\
\cmidrule(lr){2-4} \cmidrule(lr){5-7}
\textbf{Tar.} & \textbf{Pre.} & \textbf{F1 Macro} & \textbf{F1 Best} & \textbf{Pre.} & \textbf{F1 Macro} & \textbf{F1 Best} \\
\cmidrule(lr){2-4} \cmidrule(lr){5-7}
\textbf{Ar} & En & 63.63 {\small $\pm$2.71} & 65.25 & En & \textbf{67.65 {\small $\pm$0.1}} & 67.76 \\
\textbf{En} & It & 65.07 {\small $\pm$1.8} & 66.93 & Sp & \textbf{67.47 {\small $\pm$0.46}} & 67.85 \\
\textbf{Fr} & Sp & \textbf{73.55 {\small $\pm$0.92}} & 74.21 & En & 68.24 {\small $\pm$5.2} & 71.66 \\
\textbf{De} & En & 72.48 {\small $\pm$0.44} & 72.97 & It & \textbf{75.49 {\small $\pm$0.67}} & 76.18 \\
\textbf{Hi} & It & 33.57 {\small $\pm$9.34} & 39.41 & It & \textbf{55.35 {\small $\pm$0.38}} & 55.68 \\
\textbf{It} & Ge & 70.43 {\small $\pm$1.51} & 71.4 & Po & \textbf{73.5 {\small $\pm$0.58}} & 74.12 \\
\textbf{Pt} & Sp & 71.87 {\small $\pm$0.24} & 72.14 & En & \textbf{76.08 {\small $\pm$1.08}} & 76.78 \\
\textbf{Es} & Po & 67.68 {\small $\pm$0.87} & 68.66 & Po & \textbf{68.68 {\small $\pm$0.2}} & 68.85 \\
\bottomrule
\end{tabular}
}
\caption{\label{tab:cross-lingual} Bilingual experiments. We finetune XLM and XLM-Twitter models for S/A in the target language (Tar.) but instead of starting with random initialization of the adapter, we start with the adapter pretrained (Pre.) in the language that best performed in the zero shot classification for the Target language (using validation).}
\end{table}

\begin{table}[!t]
\scalebox{0.90}{ 
\begin{tabular}{clclc}
\toprule
 & \multicolumn{2}{c}{\textbf{XLM}} & \multicolumn{2}{c}{\textbf{XLM-Twitter}} \\
\cmidrule(lr){2-3} \cmidrule(lr){4-5}
 & \multicolumn{1}{c}{\textbf{F1 Avg}} & \textbf{F1 Best} & \multicolumn{1}{c}{\textbf{F1 Avg}} & \textbf{F1 Best} \\
\cmidrule(lr){2-3} \cmidrule(lr){4-5}
\textbf{Ar} & 64.31 {\small $\pm$1.92} & 66.52 & \textbf{66.89 {\small $\pm$1.18}} & 67.68 \\
\textbf{En} & 68.52 {\small $\pm$1.42} & 69.85 & \textbf{70.63 {\small $\pm$1.04}} & 71.76 \\
\textbf{Fr} & 70.52 {\small $\pm$1.76} & 72.24 & \textbf{71.18 {\small $\pm$1.06}} & 72.32 \\
\textbf{De} & 72.84 {\small $\pm$0.28} & 73.15 & \textbf{77.35 {\small $\pm$0.27}} & 77.62 \\
\textbf{Hi} & 53.39 {\small $\pm$2.00} & 54.97 & \textbf{56.39 {\small $\pm$1.60}} & 57.32 \\
\textbf{It} & 68.62 {\small $\pm$2.23} & 70.97 & \textbf{69.06 {\small $\pm$1.07}} & 70.12 \\
\textbf{Pt} & 69.79 {\small $\pm$0.57} & 70.37 & \textbf{75.42 {\small $\pm$0.49}} & 75.86 \\
\textbf{Es} & 66.03 {\small $\pm$1.31} & 66.94 & \textbf{67.91 {\small $\pm$1.43}} & 69.03 \\
\cmidrule(lr){1-5}
\multicolumn{1}{l}{\textbf{All}} & 66.93 {\small $\pm$0.16} & 67.07 & \textbf{69.45 {\small $\pm$0.63}} & 70.11 \\
\bottomrule
\end{tabular}
}
\caption{\label{tab:multilingual} Multilingual experiments. XLM-R and XLM-Twitter are finetuned using one single multilingual dataset. We evaluate the two multilingual models with the test set of each language and with the composition of all the test sets (All). F1 macro is the average of three runs and F1 best is the best one of them.} %\red{TODO: Merge with Table 4, remove F1 Best, add reference to monolingual result}}
\end{table}

\end{document}